# Drone swarms in fire suppression activities

*Elena Ausonio*[A], *Patrizia Bagnerini*[AC]*, and Marco Ghio*[B]

[A]University of Genoa (DIME), Via all'Opera Pia 15, 16145, Genoa, Italy.
[B]Inspire S.r.l. Via Marcello Durazzo 1/9, 16122, Genoa, Italy.
[C]Corresponding author. Email: bagnerini@dime.unige.it

**Abstract**. Recent huge technological development of Unmanned Aerial Vehicles (UAVs) can provide breakthrough means of fighting wildland fires. We propose an innovative forest firefighting system based on the use of a swarm of hundreds of UAVs able to generate a continuous flow of extinguishing liquid on the fire front, simulating rain effect. Automatic battery replacement and refilling of the extinguishing liquid ensure the continuity of the action, and fire-resistant materials protect drones exposed to possible high temperatures. We demonstrate the validity of the approach in Mediterranean scrub first computing the critical water flow rate according to the main factors involved in the evolution of a fire, then estimating the number of linear meters of active fire front that can be extinguished depending on the number of drones available and the amount of extinguishing fluid carried. A fire propagation cellular automata model is also employed to study the evolution of the fire. Simulation results suggest that the proposed system can successfully integrate, or in case of low-intensity and limited extent fires completely replace, current forest firefighting techniques.

**Introduction**

Forest environments, woods and green areas constitute a natural resource for human life. Unfortunately, they are progressively risking impoverishment and destruction [1,2]. Wildfires represent the most serious and prevalent threat to Mediterranean forests, causing erosion and chemical alteration of the soil, and climate changes [3,4]; they are a particularly complex phenomenon, influenced by numerous interdependent variables, some of which are constantly evolving in time. In the centuries, slash-and-burn practices have been used by humans for land management to adapt the territories for agricultural and residential activities [5]. In the past few decades, there has been a countertrend to an ever-increasing abandonment of rural lands that has led to a re-colonization of many woodlands and shrublands. These richer rural landscape conditions have caused an increase of fire hazard, of wildfires frequency, incidence and magnitude, and dangerous consequences involving people, things, and natural environment; as a result, the issue of forest firefighting has become increasingly important [6].

The protection of forests is implemented through the management of wildfires following a three-step process: prevention, detection and suppression [7], and involves fire brigade, civil protection, police forces, and volunteers. Prevention activity consists in the execution of all actions aimed at mitigating the risk of fire and its consequent damages, in order to make the natural contexts safer and to achieve the objectives of ecosystem management. Prevention interventions, aimed at reducing the potential causes of fire ignition, may consist of: preventive silviculture techniques, prescribed fire, interventions on firebreaks, and access roads to the forest [8].

The suppression phase includes both direct and indirect attack [9]. Generally, the first strategic approach is the direct attack, usually chosen for small or large fires of low-to-medium intensity. It consists in actions performed directly against the flames at the immediate edge of the fire front or strictly parallel to it; it is commonly associated with the use of water as the main extinguishing component. In emergencies, direct attack may also involve cooling with chemical additives mixed with water to increase suppression effectiveness. Retardant additives are inorganic salts, mainly ammonium phosphates, which inhibit the flame combustion and can slow down fire progression when the water used for their diffusion has evaporated [10]. Ground suppression forces are equipped with different vehicles and firefighting systems, e.g., pumps, bulldozers, tankers with different water capacities, which are employed depending on the nature of the fire [7,11].

Aerial support is another important firefighting tool capable of providing an increasing amount of extinguishing liquid. Among the types of aircraft are helicopters, single-engine tankers, fixed-wing water-scooping aircraft, and large multi-engine tankers. These aircrafts differ in cost, flight speed, flight distance, response time, maneuverability, tank capacity, and type and effectiveness of the liquid, e.g., water, suppressant or retardant, that can be delivered [12]. In case of significant water sources in the surrounding area, planes and helicopters can collect water and drop it on both the front and the central part of the fire to counteract its evolution. As a result, airborne resources have enhanced the effectiveness of the initial attack, as they reach the fires rapidly, even in inaccessible locations. As a consequence of the proximity to the sea, tanker planes (e.g., Canadair) and helicopters are widely used in Mediterranean European countries [3]. Aviation fire resources are able to create containment lines before the arrival of ground crews and also provide punctual protection for structures and other threatened assets.

The main disadvantage of direct attack is the risk to human life due to the short distance from the fire: firefighters risk entrapments [13] and are exposed to heat, smoke and flames during operations when conducting fire control actions directly adjacent to the edge of the fire [14,15]. Drawbacks are also present with regard to the use of firefighting aviation: for safety reasons, in certain countries, Canadair can only be carried out during the day and allows to perform a limited number of drops, given the need after each intervention to refuel at an appropriate nearby place. Furthermore, the costs of these strategies have been examined in literature. Aerial firefighting has a high cost of purchasing, operating and maintaining aircraft, and staff training [12,16].

When the intensity of a fire makes direct attack unsafe and wildfire behavior exceeds the firefighter's capacity of extinction, a proactive approach is needed. Indirect attack consists in anticipating the flames, building impermeable firelines along a predetermined route at varying safe separation distances from the advancing fire. Advanced firefighting organizations plan extinction operations, identifying appropriate and safety areas where they can organize indirect attack: build effective and distant fire control lines with burnout or backburn actions [17].

As previously mentioned, a second phase should follow prevention and precede suppression with direct and indirect attacks: the detection phase, i.e., the identification and monitoring of an actual fire in progress [18,19]. A fire of limited size is known to be easier to control and extinguish than a fire that has already spread, hence, for a faster detection, valid tools are the Unmanned Aerial Vehicles (UAVs).

During the twentieth century, the UAVs, also generically called drones, were exclusively used for military applications, whereas in the last few years, their use in civil applications has increased. Both research institutions and universities, as well as industry, have shown a growing interest in studying this technology. Promising applications include surveillance, precision farming, inspection of potentially dangerous sites, and environmental monitoring.

The use of UAVs has raised in forest fire prevention and detection as well, also thanks to techniques such as image and three-dimensional point cloud processing [7,20–24]. In the United States, a law enacted in 2019, the "John D. Dingell, Jr. Conservation, Management, and Recreation Act", pushes federal agencies to expand the use of drones in managing and fighting wildfires [25]. Small unmanned aerial systems are generally employed in three stages: fire search, fire confirmation, and fire observation [26]. They have a short reaction time and flexibility of use, and they reach locations that are inaccessible and dangerous for humans. In addition, thanks to their view from above using video or thermal cameras, the UAVs are able to help fire brigade in reconstructing overviews of an incident [27] and they can provide data on chemical measurements, when equipped with appropriate technical sensors. Civil protection and firefighters equip themselves with drones to track vegetation and areas at risk and to carry out evaluations and interventions based on images and data transmitted in real time by drones flying over the burning area. The main limitations of this use are the difficulty of going beyond a sporadic monitoring that does not allow a continuity in time (both the phase of data acquisition and processing must be done manually by an operator) and the impossibility of acting promptly to alert in case of an outbreak of fire. Generally, the overflight of drones takes place with the fire in progress to estimate the extent and direction of propagation of the flames. Recent technological advances in UAV field, have increased the possibility to provide real time and high-quality information to end-users. On this issue, in [28], an algorithm for estimating the fire propagation is proposed with the aim of enabling intelligent and long-lasting coordination of UAVs to support firefighters. In [26], the authors show how multiple aerial vehicles with on-board infrared or visual camera, can collaborate to automatically obtain information about the evolution of the fire front shape and other parameters.

The system in [29] is developed to help fire experts in indirect attack, by planning flights to reach previously inaccessible terrain and deliver ignition spheres to light prescribed fires.

Unfortunately, there are still few studies in the literature that suggest the employment of UAVs not only for the prevention and monitoring of forest fires, but even with the aim of extinguishing them [19,30]. For instance, [31,32] and [33] promote the idea of using such technology in firefighting applications, especially in areas difficult to reach by humans. The first propose a fleet of self-organized drones, carrying 120 $kg$ of water each, with a coordination mechanism based on a forgetful particle swarm algorithm; the second present a rotary wing UAV equipped with a payload drop mechanism that can carry firefighting spheres and release them against fires. [34] proposes the use of drones carrying extinguishing balls as a supplement to traditional firefighting methods. In [35], an aerial robot made of fire-resistant materials is introduced to avoid the risk of damage to electronic equipment directly exposed to flames.

In this work, we propose an innovative method based on the use of a swarm of collaborative UAVs able to transport large quantities of fractionated extinguishing liquid and to release it on fire fronts, simulating rain effect (see Fig. 1). This system is designed not only for the detection of a fire, but also for its suppression, in cooperation with the resources previously introduced, i.e., aerial forces and ground crews, with the aim of reducing the risk for human life. We first introduce the drone system that manages the swarm of UAVs. Then, we demonstrate its effectiveness showing that the amount of extinguishing liquid is adequate in suppressing or at least containing a low-intensity wildland fire. For this purpose, the critical water flow rate, i.e., the rate of water application required to arrest a certain number of linear meters of active fire front, is estimated based on fire parameters such as flame length, wind speed, humidity and height of vegetation. We have chosen to use plant species typically present in the Mediterranean scrubland, which do not give rise to flames of excessive height. Therefore, it is not strictly necessary to take into account the percentage of water that cannot reach the fire due to the foliage of the trees. Then, assumed that the drone system is positioned at a certain distance from the fire depending on various surrounding conditions, the number of drones necessary to assure such flow of water is computed. Finally, the implementation of a fire cellular automata propagation model allows to predict how the fire front subjected to the action of drones is modified over time.

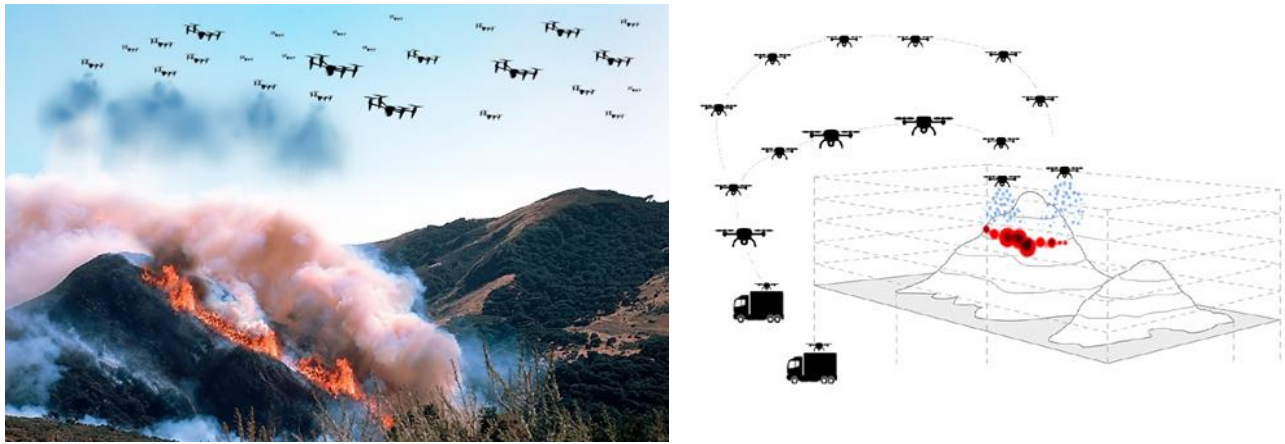

**Fig. 1.** Representation of the proposed firefighting system based on the use of a swarm of hundreds of collaborative UAVs.

*Description of the drone swarm management system*

The management system of the drone swarm has to meet a number of requirements in order to be used effectively in firefighting: (i) it has to ensure operational continuity without downtime for a high number of hours (virtually also H24); (ii) it has to be able to automatically replace exhausted batteries and insert it into a charging circuit, given the limited battery duration of the current drones (just over half an hour); (iii) it has to allow for multiple refills of the drones with an appropriate extinguishing liquid placed in containers docked to drones. It mainly consists of a support unit that manages the drone swarm (henceforth called platform) that can be easily moved and positioned close to the fire front. In addition, drones exposed to possible high temperatures have to be assembled with fire-resistant materials.

A system that satisfies these requirements has considerable advantages in fighting forest fires: it can be used both during the day and at night and in low visibility conditions, unlike common aircraft; it does not require the presence of an available water basin in the vicinity; it can be used continuously until the alarm ceases, since the automatic battery change, the continuous charging system and the complete automation of the payload switch (containing the extinguishing liquid), ensures virtually a H24 duration; it can be used in areas not directly accessible by firefighting equipment and in areas with complex orography; it is a precision system because the flight plan, the area concerned, etc. can be planned in an accurate manner; it is a flexible system in which the area of intervention can be modified in real time as the conditions of the fire evolve; it does not put at risk the lives of pilots of aircraft that often have to operate in conditions of considerable danger.

The amount of water transported by a drone is obviously significantly reduced compared to the volume of water of a firefighting aircraft [36]. At the same time, however, a large number of drones can guarantee a temporal continuity and uniformity of diffusion that an aircraft is unable to obtain due to the timing of provision. One of the authors of this paper is the inventor of the patent "Methods and apparatus for the employment of drones in firefighting activities" [37] that precisely concerns the creation of a system for drones able to drop small quantities of firefighting liquid on wildland fires.

Multiple strategies are possible: either directing the action of drones directly on the flames, or on an adjacent area, so as to increase the humidity of the vegetation and prevent the spread of flames to particularly sensitive areas, for example occupied by inhabited areas or installations at high risk. Moreover, the drone system creates the so-called rain effect, i.e., dropping small quantities of firefighting liquid by drizzling it over the fire [38] or the surrounding vegetation, instead of spreading it in a concentrated manner. This method, both theoretically and experimentally, is acknowledged as being particularly effective in domestic and/or industrial firefighting systems (fire sprinkler systems) [39]. Although it is not the purpose of this work to assess this effect by comparing it with the impact produced by aircraft carrying the same amount of water, it is an interesting feature that requires further investigation. At the moment, the cost of the system compared to the current firefighting tools cannot be estimated, as it depends on many factors, primarily the implementation choices and the cost of drones. However, the price of UAVs has been continuously decreasing in recent years thanks to significant technological advancement in the field, with costs expected to decrease further.

The drone system involves the use of a platform that manages drones and changes their battery and payload. Such a platform can be considered as a base station for drones. Drone base stations are gaining popularity mainly for their use in video surveillance and inspections [40]. In the field of forest firefighting, there is very little literature concerning the use of drone base stations. Among them we mention the charging base designed in [24] for fire surveillance, equipped with an upper sliding door and a vertically moving bed for take-off and landing maneuvers. The aim of this work is to prove that a management system of a drone swarm that fulfils the requirements (i)-(iii) is able to generate a sufficient flow of extinguishing liquid to effectively fight a low-intensity forest fire. Moreover, the solution could be perfectly applicable in case of hotspots, i.e., pre-fire and post-fire outbreaks and burning embers, identifying these areas with drones carrying thermal cameras. These situations generally do not require the intervention of air tankers, and in the meantime the hotspots are often difficult to reach by fire brigades.

Technical implementations of the system and the management of a swarm of drones on the fire front will be analyzed in future papers, but particular care has been taken in choosing reasonable ranges of parameters concerning the amount of extinguishing liquid carried by each drone [41], the timing of automatic battery and payload replacement and the number of drones that a platform can handle in an interval of time.

**Calculation of critical water flow rate CF**

The main substance used to suppress wildfires is water, thanks to its high availability, low cost and great extinguishing capacity [42]. In order to test the validity and effectiveness of using a large number of drones to contain and extinguish forest fires, it is essential to estimate the critical water flow rate (CF). CF is the rate of water application required to arrest a certain number of linear meters of active fire front or, if the water is sufficient, to completely extinguish a wildfire. Once this value has been determined according to the different fire parameters, it is then feasible to compute the number of drones required, knowing the amount of water carried by each drone. It is also possible to reverse the calculation, i.e., to compute the number of active fire front meters contained by a given number of drones and/or platforms.

The literature contains a number of works predicting $CF$ to extinguish a wildland fire within an infinite period of time available [43,44]. These articles are based on models obtained by physical considerations or by experimental data collected in databases of thousands of fires.

In this paper, we use Hansen's approach [43] based on the Fire Point theory [45,46] in which the energy balance at the fuel surface when water is applied is given by

$$(f\,\Delta H_c - L_v)\,\dot{m}''_{cr} + \dot{q}''_E - \dot{q}''_L - \dot{q}''_{water} = 0 \tag{1}$$

where $f$ is the heat release fraction transferred back to the fuel surface by convection and radiation, $\Delta H_c$ is the effective heat of combustion ($kJ\,kg^{-1}$), $L_v$ is the heat of gasification of the fuel ($kJ\,kg^{-1}$), $\dot{m}''_{cr}$ is the mass burning rate per unit fuel area at the critical point ($kg\,m^{-2}\,s^{-1}$), $\dot{q}''_E$ is the external heat flux ($kW\,m^{-2}$), $\dot{q}''_L$ and $\dot{q}''_{water}$ are the heat loss from the surface and due to the water vaporization ($kW\,m^{-2}$), respectively.

Using Spalding's B Number theory [47], $\dot{m}''_{cr}$ is computed as

$$\dot{m}''_{cr} = \frac{h}{c_p}\,\ln\left(1 + \frac{Y_{O_2}\,\Delta H_{R,O_2}}{\phi\,\Delta H_c}\right) \tag{2}$$

where $h$ is the convective heat transfer coefficient ($kW\,m^{-2}$), $c_p$ is the specific heat of air at constant pressure ($kJ\,kg^{-1}\,K^{-1}$), $Y_{O_2}$ the oxygen mass fraction, $\Delta H_{R,O_2}$ the heat of combustion per unit mass of oxygen consumed ($kJ\,kg^{-1}$), and $\phi$ is the fractional convective heat loss from the flame required for quenching the flame.

The heat loss due to water vaporization is given by [43]:

$$\dot{q}''_{water} = \eta_{water}\,\dot{m}''_{water,cr}\,L_{v,water} \tag{3}$$

where $\eta_{water}$ is the efficiency of water application, $L_{v,water}$ is the enthalpy change of water at 283 K and water vapour at 373 K ($kJ\,kg^{-1}$), and $\dot{m}''_{water,cr}$ is the critical water application rate ($kg\,m^{-2}\,s^{-1}$). Then, replacing (3) in (1) and $f$ by $\phi$, the critical flow rate CF is expressed as

$$CF := \dot{m}''_{water,cr} = \frac{1}{\eta_{water}\,L_{v,water}}\left((\phi\,\Delta H_c - L_v)\,\dot{m}''_{cr} + \dot{q}''_E - \dot{q}''_L\right) \tag{4}$$

The expression of $\dot{q}''_E$ is given by

$$\dot{q}''_E = \dot{q}''_{E,rad} + \dot{q}''_{E,conv} \tag{5}$$

The first term $\dot{q}''_{E,rad}$ is the incident flame radiation per unit area and is given by [43]

$$\dot{q}''_E = \frac{r_c\,I}{2\,L_f + D}\,\phi\,\tau \tag{6}$$

where $r_c$ is the radiative component per unit length of fire front, $I$ is the fireline intensity, i.e., the heat release rate per unit length of fire front, $L_f$ the flame length, $D$ the depth of active combustion zone ($m$), and $\tau$ the atmospheric transmissivity. The second term $\dot{q}''_{E,conv}$ that represents the convective heat transfer has to take into account in the case of large wind speed, high-intensity fires or steep terrain [43]. Since here low-intensity fires and flat ground are considered, $\dot{q}''_{E,conv}$ is different from zero only when the flame angle $A$, i.e., the angle between the flame and the unburned fuel ahead, is lower than 30° [43]. Following [48,49,43], $A$ is computed by numerically solving the following system of two nonlinear equations

$$\begin{cases} \tan(90 - A) - 1.22\sqrt{U^2/(g\,H_f)} = 0 \\ H_f - L_f \sin A = 0 \end{cases} \quad (7)$$

where $H_f$ (m) is the flame tip height, $U$ the mean horizontal wind speed and $g$ is the gravity acceleration ($m\,s^{-2}$). Therefore, the expression of $\dot{q}''_{E,conv}$ is given by

$$\dot{q}''_{E,conv} = \begin{cases} h\,(T_g - T_{fuel}) & if\ A < 30° \\ 0 & if\ A \geq 30° \end{cases} \quad (8)$$

where $T_g$, $T_{fuel}$ are gas and fuel surface temperature, respectively ($K$).
The heat loss from the surface $\dot{q}''_L$ is given

$$\dot{q}''_L = \dot{q}''_{L,rad} + \dot{q}''_{L,conv} \quad (9)$$

where radiative $\dot{q}''_{L,rad}$ heat loss is

$$\dot{q}''_{L,rad} = \varepsilon\,\sigma\,(T^4_{fuel} - T^4_a) \quad (10)$$

and convective $\dot{q}''_{L,conv}$ heat loss is

$$\dot{q}''_{L,conv} = h\,(T_{fuel} - T_a) \quad (11)$$

The coefficient $\varepsilon$ represents the fuel emissivity, $\sigma$ the Stefan-Boltzmann constant ($W\,m^{-2}\,K^{-4}$), $T_a$ ($K$) the ambient temperature. Finally, using Byram relation [43], fireline intensity $I$ can be expressed as a function of flame length $L_f$, which corresponds in low-intensity surface fires to

$$I = 259.833\,L_f^{2.174} \quad (12)$$

It is therefore possible to calculate the critical flow rate as the flame length varies, using equation (4). The computation of CF described in this section is valid for various fuel models, but some of the coefficients depend on the vegetation chosen. Table 1 shows the coefficients used with their bibliographic reference. For parameters depending on the type of vegetation was selected the Mediterranean scrub, while in Hansen's paper [43] coefficients are referred to Pinus pinaster. In Fig. 2, the critical flow rate at low and high intensity and wind speed at 0 and 10 $m\,s^{-1}$ is displayed.

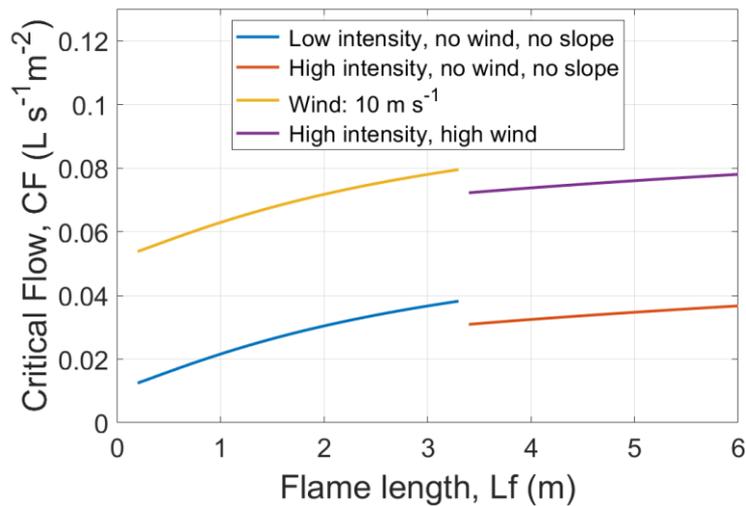

Fig. 2. Critical water application rate, as a function of the flame length, for various conditions. The heat release rate per unit area in the active combustion zone ($I_r = I\,D^{-1}$) was assumed to be steady-state at 500 $kW\,m^{-2}$ on the left and 2000 $kW\,m^{-2}$ on the right.

**Table 1. CF parameters**
Parameters used to compute the critical water flow rate

| Symbol | Parameter | Value | References |
|---|---|---|---|
| $\Delta H_c$ | effective heat of combustion | $19500\ kJ\ kg^{-1}$ | [50] |
| $L_v$ | heat of gasification of the fuel | $1800\ kJ\ kg^{-1}$ | [43] |
| $h$ | convective heat transfer coefficient | $20\ kW\ m^{-2}$ | [43] |
| $c_p$ | specific heat of air at constant pressure | $1\ kJ\ kg^{-1}\ K^{-1}$ | [43] |
| $Y_{O_2}$ | oxygen mass fraction | 0.233 | [43] |
| $\Delta H_{R,O_2}$ | heat of combustion per unit mass of oxygen consumed (Genista salzmannii) | $13480\ kJ\ kg^{-1}$ | [51] |
| $\phi$ | fractional convective heat loss | 0.3 | [43] |
| $\eta_{water}$ | efficiency of water application | 0.7 | [43] |
| $L_{v,water}$ | enthalpy change of water | $2640\ kJ\ kg^{-1}$ | [43] |
| $\tau$ | atmospheric transmissivity | 1 | [43] |
| $r_c$ | radiative component per unit length of fire front (Erica arborea) | 0.20 | [52] |
| $\varepsilon$ | fuel emissivity | 0.6 | [43] |
| $\sigma$ | Stefan-Boltzmann constant | $5.67 \cdot W\ m^{-2}\ K^{-4}$ | [43] |
| $T_{fuel}$ | fuel surface temperature (Cistus monspeliensis) | $693\ K$ | [53] |
| $T_g$ | gas temperature | $800\ K$ | [54] |
| $T_a$ | ambient temperature | $293\ K$ | [43] |
| $W$ | total fuel load | $15\ t ha^{-1}$ | [44] |

In the following, we calculate the variation of CF as a function of changes in wind speed and moisture content. In that case, it is possible to replace equation (12) for fireline intensity, with the formula based on the rate of spread

$$I = \frac{\Delta H_c\ W\ RoS}{36} \quad (13)$$

where $W$ is total fuel load considering fine fuels typically less than 6 mm in diameter and $Ros$ is the rate of spread of the fire front. Among the various types of $RoS$ was chosen Fernandes' rate of spread [55], which is suitable for low-intensity fires:

$$RoS = 0.06\ a\ U^b \exp(-cM_d) \quad (14)$$

where $U$ is the wind speed measured at $2\ m$ above ground level in $km\ h^{-1}$, $M_d$ is the moisture content percentage of the elevated dead fuels, and $a$, $b$, and $c$ are some parameters obtained experimentally by non-linear regression analysis for Mediterranean scrub (see Table 2). Differently from [55], the rate is multiplied here by 0.06 to be expressed in $km\ h^{-1}$. Using equations (4) and (13), $CF\ (L\ min^{-1})$ required to extinguish $1\ m$ section of active head fire front is obtained as a function of wind speed $U$, moisture content $M_d$, and active flame depth $D$. Fig. 3 shows some graphs of $CF$ by varying two of the three factors, maintaining the third one fixed. As expected, $CF$ is directly proportional to wind speed and active flame depth and inversely proportional to moisture content. The discontinuities in the curves are due to the computation of $\dot{q}''_{E,conv}$ as the angle $A$ varies in equation (8).

**Impact of drones on the evolution of the active fire front**

Based on the critical water flow rate computed in the previous section, it is possible to estimate the number of linear meters of active fire front that can be extinguished as these factors vary. The platform is assumed to be positioned at a certain distance from the fire depending on various boundary conditions, such as wind direction, terrain orography, presence of roads. A certain time interval is also required to automatically switch the battery and the payload (the extinguishing liquid) carried by each drone; this timespan can vary from a minimum of a few seconds to a maximum of one minute. A very short, almost instantaneous, time is also required for the liquid to be released on the fire by each drone. Since in this work we do not describe how these technical features would be effectively implemented, we consider them in a single variable: the time interval $\Delta t$ ($min$) in which a drone arrives on the platform, is charged with a new payload, takes off, reaches the active fire front, releases the liquid and lands back on the platform. A $\Delta t$ equal to 6 minutes is assumed to remain fairly cautious, but even lower values are possible if the platform is positioned against the wind next to the active fire front, as in the case of firefighters' vehicles.

Table 2. Critical flow rate parameters
The parameters have been computed in (Fernandes 2001)

| Parameter | Symbol | Value |
|---|---|---|
| Rate of spread parameters | a | 3.258 |
|  | b | 0.958 |
|  | c | 0.111 |
| Active flame depth | D | 2 (m) |

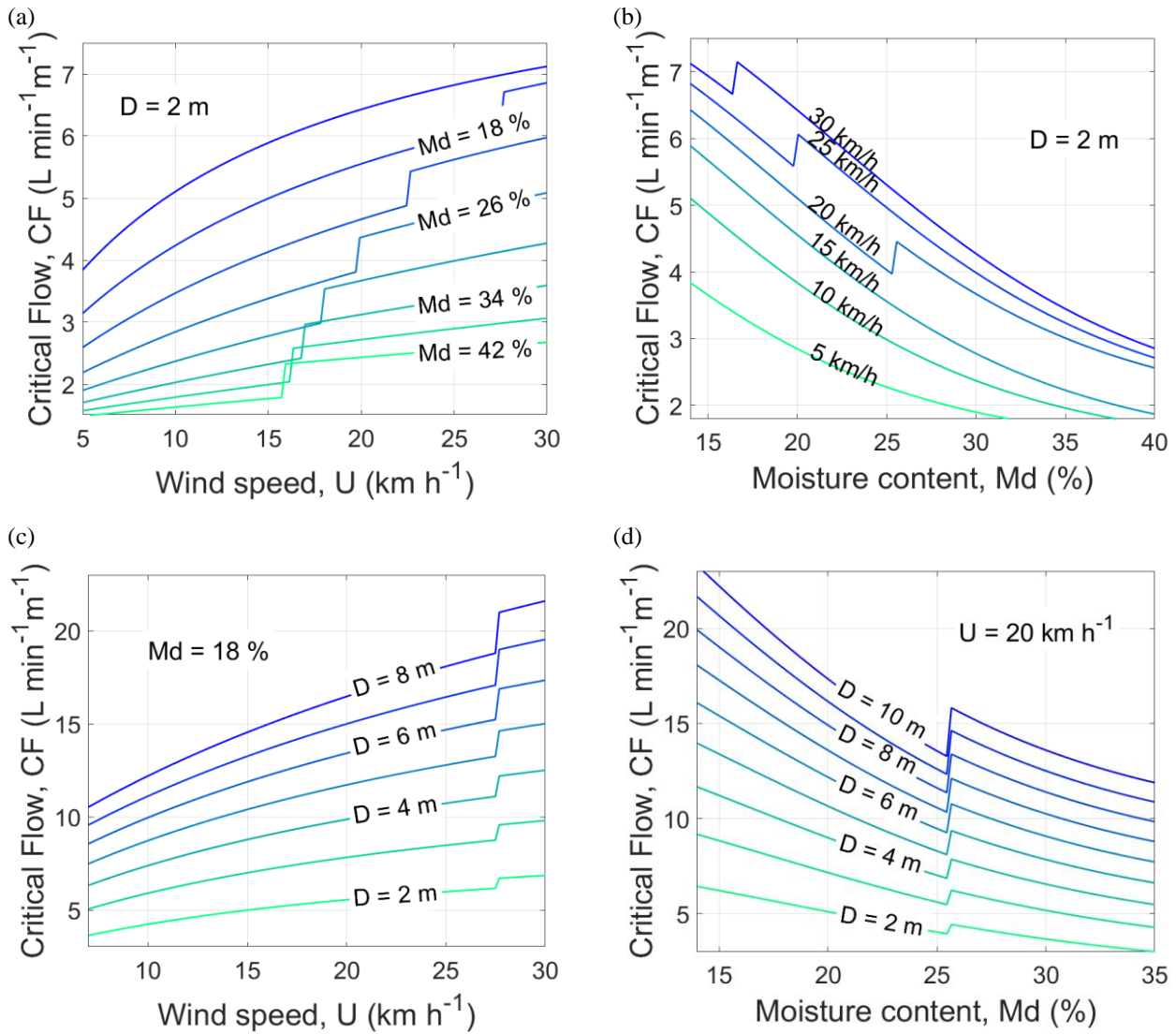

Fig. 3. Critical water flow CF computed as a function of the main parameters of the fire. (*a*, *b*) $CF$ with wind speed and moisture content varying, respectively, and $D$ fixed at 2 $m$. (*c*) $CF$ as a function of wind speed with $D$ varying in the different curves and $M_d$ fixed to 18%. (*d*) $CF$ as a function of moisture content while $D$ varying in the different curves and wind fixed at 20 $km\ h^{-1}$.

The liquid carried by each drone is identified by $L_d$ ($L$). It is reasonable to assume this value to be between 10 and 50 $L$. In fact, the coordinated system proposed concerns the use of a large number of small-sized drones carrying a limited amount of liquid, in contrast to self-guided vehicles similar to current firefighting aircraft with a capacity of thousands of liters. Moreover, each platform is capable of handling a number $n_d$ of drones, ensuring battery replacement and liquid refilling for each drone. Reasonable values of $n_d$ are between 80 and 120. Once these variables are introduced, it is easy to estimate the water flow that a platform assure in the time unit. Each drone is able to deliver $n_h$ discharges of extinguishing liquid per hour corresponding to

$$n_h = \frac{60 \ (min)}{\Delta t \ (min)} \tag{15}$$

Therefore, a platform, managing $n_d$ drones, is designed to deliver $n_h^{tot}$ discharges of extinguishing liquid per hour

$$n_h^{tot} = n_h \ n_d \tag{16}$$

and to spread $L_h^{tot}$ of liquid equals to

$$L_h^{tot} = L_d \ n_h^{tot} \ (L \ h^{-1}) \tag{17}$$

Dividing this value for 60 (minutes), it is possible to estimate the drones flow rate ($DF$), i.e., the amount of liquid that one platform spreads each minute

$$DF = \frac{L_h^{tot}}{60} \ (L \ min^{-1}) \tag{18}$$

Replacing in (18) the expression of the variables stated in (15) - (17), we obtain the flow rate that the platform handling $n_d$ drones spreads per minute:

$$DF = \frac{L_d \ n_d}{\Delta t} \ (L \ min^{-1}) \tag{19}$$

For instance, a platform with $n_d = 120$ drones, each carrying $L_d = 20$ liters of fire extinguishing liquid and completing a round trip in $\Delta t = 6$ min, ensures a flow of 400 ($L \ min^{-1}$). Once calculated the continuous flow that a platform can guarantee and the flow rate necessary to extinguish one linear meter of active front, it is possible to estimate the number $m_f$ of linear meters that can be extinguished:

$$m_f = \frac{DF}{CF} \ (m) \tag{20}$$

or equivalently

$$m_f = \frac{L_d \ n_d}{\Delta t \ CF} \ (m) \tag{21}$$

where $CF$ is computed using equation (4). Since the calculation of linear meters does not take into account the time required to ensure the flow of extinguishing liquid, drones are assumed to continue to provide such flow throughout the duration of the fire. Fig. 4 shows the linear meters of fire that can be arrested by using the proposed firefighting method. For example, approximately 70-75 linear meters of active front can be extinguished with 120 drones each carrying 20 $L$ or with 80 drones carrying 30 $L$ (Fig. 4a). These results show that a platform managing up to 120 drones is a valid alternative to current firefighting systems in the case of small or low-intensity fires. In a large wildland fire, the system can control a part of the front, e.g., to prevent advance in critical areas. The effectiveness can be further improved by the simultaneous use of multiple platforms that can attack the fire front from multiple sides. Moreover, the effect of the platform can be directed to the fire front or to areas that have not yet caught fire thereby creating a firebreak without risk for firefighters who are not forced to approach the fire.

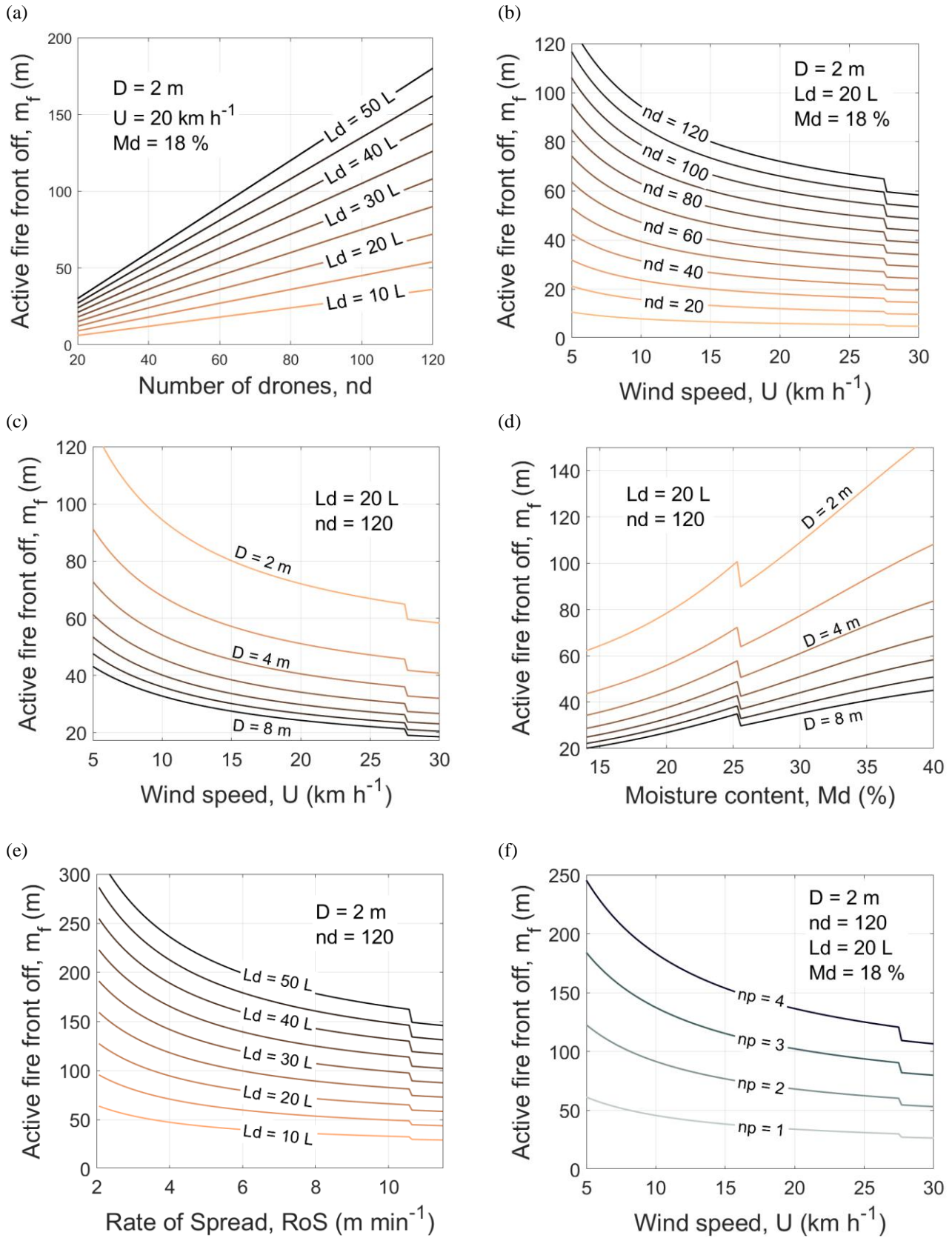

**Fig. 4.** Linear meters of fire $m_f$ that can be arrested by using the proposed firefighting method. In figures (*a, b, c, d*) the active flame $D$ is fixed to 2 $m$. (*a*) $m_f$ as a function of the number of drones with the extinguishing liquid carried by drones varying in the curves, wind speed and moisture content are fixed to 20 $km\ h^{-1}$ and 18%, respectively. (*b*) $m_f$ as a function of wind speed with the number of drones varying in the curves. The moisture content is 18% and the liquid carried by drones is 20 $L$. (*c, d*) $m_f$ as a function of wind speed and moisture content, respectively, with active flame $D$ varying in the curves. Drones are 120, each one carrying 20 $L$. (*e*) $m_f$ as a function of the rate of spread with liquid carried by drones varying in the curves. The number of drones is 120 . (*f*) $m_f$ as a function of wind speed with the number of available platforms varying in the curves. Each platform manages 120 drones each of them carrying 20 $L$. The moisture content is 18%.

**Estimate of the drones required to extinguish a specified number of linear meters of active fire front**

Considering the total or partial extent of the fire front whose propagation is to be prevented, it is possible to determine the number of drones required. It is assumed to have drones carrying $L_d$ liters of extinguishing liquid and to locate the platform so that the time interval required for a drone to reach the fire, release the liquid onto it and return to the platform is $\Delta t$ ($min$). Depending on the fire parameters, wind, moisture content, etc., the requested water flow rate changes and consequently the required number of drones. The drone flow rate $DF_r$ required to extinguish $m_r$ of active fire front is

$$DF_r = m_r\, CF \quad (L\, min^{-1}) \tag{22}$$

since $CF$ is the flow rate for 1 linear meter. Therefore, replacing (22) in (19), we obtain the number $n_d$ of drones required:

$$n_r = \frac{CF\, \Delta t\, m_r}{L_d} \tag{23}$$

For instance, in a fire with a moisture content of 18% and a wind speed of $10\, km\, h^{-1}$, to extinguish 70 meters of active front with drones carrying 30 liters and completing a loop in $\Delta t = 6\, min$, the number of drones required is 60.

**Cellular Automata model for studying the effect of the UAV platform on fire evolution**

In the previous sections, we estimated the number of linear meters of active front arrested using one or more platforms handling a given number of drones carrying extinguishing liquid. Starting from this value, in this section we use a fire propagation model for studying the evolution of a fire as a result of the containment action of the platforms. Forest fires are a particularly complex phenomenon, influenced by numerous interdependent variables, some of which are constantly evolving in time. Risk assessment, propagation and effect models are the three categories in which fire models are grouped by [56]. The objective of forest fire simulation is therefore to improve prevention and control operations: assessment of the attack surface, prediction of the evolution of the fire front, preventive mobilization of rescue teams, and containment of the front-line and fire extinction. In our case, in order to prove the effectiveness of using a coordinated system of drones, among all models of fire evolution, we choose a cellular automata model to simulate and calculate the modification of the front by the contribution of the extinguishing liquid provided by drones.

Cellular automata (CA) are mathematical idealizations of physical systems, represented by connected and organized elements that interact with each other and constitute a single entity with the external world. The definition of the physical environment determines the universe upon which the CA is modelled, physical quantities take on a finite set of discrete values [57], depicted in grids (2 or 3 dimensional lattices) that evolve at discrete time intervals, according to stochastic rules. Every single cell has a finite state characterized by one or more variables and the respective numerical values. Cell states vary according to a local transition function applied to all cells in the lattice, updated synchronously and simultaneously. Specifically, the state of a cell $(i,j)$, at a given time $t$, depends only on a transition function and on the state of the cell itself and of neighboring ones at the previous discrete time step.

CA have proven their strength in predicting macroscopic and complex dynamics using simple rules that define the physics of a phenomenon on a microscopic grid scale. For this reason, CA models emerge as a useful choice for modelling the complex behavior of wildfire spread [58]. In several researches, CA have been applied to simulate fire spread for the purpose of assisting firefighters in identifying fire suppression tactics and in planning policies for fire risk management [59,60]. They can also be easily integrated with digital data from Geographic Information Systems or other sources including local meteorological data [61–63].

CA can be identified by the geometry of the regular cell arrangement, i.e., square or hexagonal cells in two-dimensional case [64,65], and the number of neighboring cells taken into account: 4 neighbors in the case of the Von Neumann neighborhood, 8 neighbors in the Moore neighborhood [66]. In [67], the authors present a novel algorithm for wildfire simulation through CA, which is able to effectively mitigate the problem of distorted fire shapes, allowing spread directions that are not constrained to the few angles imposed by the lattice of cells and the neighborhood size.

In the present paper, the CA model introduced in [68,69] is utilized to simulate the evolution and the consequent confinement of a wildfire thanks to the action of one or more platforms of drones. It consists in a square-meshed grid represented as a two-dimensional matrix, easily simulating a forest area. Each cell is generally defined by a finite number of evolving states. Four states characterize the system:

- $State = 0$. The cell cannot catch fire (empty cell). This state could describe cells corresponding to parts of the territory in which there is no vegetation that can burn.
- $State = 1$. The cell contains live fuel, not yet burned (tree cell).
- $State = 2$. The cell contains material that is burning (burning cell).
- $State = 3$. The cell contains completely burned fuel (burned cell).
- $State = 4$. The cell has a continuous flow of water that provides fire extinction (see CF) thanks to the drones.

Each cell is subject to local rules that guide the evolution of the spread of the fire. At each discrete time step $t$ of the simulation, the following rules are applied to elements $(i, j)$ of the state matrix (and therefore to all cells):

- *Rule* 1 states that an empty cell $(i, j, t)$ maintains the same state without burning at next time step.
- *Rule* 2 states that if a cell contains vegetation fuel and there was at least one neighboring cell burning at the previous time step such that $(i \pm 1, j \pm 1, t - 1) = 2$, it can catch fire with a probability $P_{burn}$ greater than a certain threshold. As the wind speed increases, we also consider next-nearest cells as in [70] and in [71]. In particular, we add two layers of cells for wind at $25\ km\ h^{-1}$ and three for wind at $35\ km\ h^{-1}$.
- *Rule* 3 determines that a cell that is burning at the present moment will be completely burned at the next one. In subsequent times, it will no longer be able to spread the fire.
- *Rule* 4 implies that a previously burned cell remains burned.

Due to the square grid based on Moore neighborhood, fire can spread to the eight adjacent cells, i.e., horizontally, perpendicularly and diagonally.

In the following, all the probabilities are computed as in [68,69]. The rule 2 implies that when a cell ignites at the current time, the next instant the fire may spread to nearby cells containing unburned fuel with a $P_{burn}$ probability:

$$p_{burn} = p_0(1 + p_{veg})(1 + p_{den})\, p_w\, p_s\, p_m \qquad (24)$$

It is a function of several variables that affect fire propagation, such as fuel properties, wind conditions and topography. The probability $p_0$ measures the chance for a cell in the neighborhood of a burning one to catch fire, supposing flat terrain and no wind conditions. The other probability factors are related to the vegetation typology, to the density and humidity of fuel in each single cell, to the wind blowing over the total area, and to the landscape altitude. Vegetation is considered as a combustible material composed of a set of solid particles distributed in the environment; a density, a typology and a percentage of humidity characterize it. Three density categories are present in the model, sparse, normal, and dense, and each of them corresponds to a $p_{den}$ value. Two types of fuel were chosen, grassland and shrubland, corresponding to the typical vegetative plants of the Mediterranean environment.

The effect due to the moisture content of vegetation is calculated adopting the formulation given in [55]. It links the rate of spread to the moisture content $M_d$ and to the two coefficients, $b$ and $c$ (see Table 2), determined by regressive analysis from experimental data. In experimental studies, this last factor is determined by weighing samples of vegetation before and after drying them, which means that the formula depends on the type of vegetation. The wind-effect probability $p_w$ takes into account both wind speed and direction and is calculated using the following empirical relation

$$p_w = \exp(c_1 V) \exp\left(V c_2 (\cos\theta - 1)\right) \qquad (25)$$

where $\theta$ is the angle between the spreading direction of the fire and the direction of the wind, $c_1$ and $c_2$ are constant values. The probability related to the effect of ground elevation is a function of a parameter derived from experimental data and of the slope angle $\theta_s$:

$$p_s = \exp(a_s \theta_s) \tag{26}$$

where $\theta_s$ is calculated using

$$\theta_s = \tan^{-1}\left[\frac{E_1 - E_2}{D}\right] \tag{27}$$

The value of $D$ is taken equal to $L$ or $\sqrt{2}L$ depending on whether the cell being considered is adjacent or diagonally located to the burning cell.

The model described above has been applied to simulate the forest fire spread in order to enable for the intervention of drones to entirely or just partially suppress the fire. The environment is completely simulated, and it is not based on a real case study. The territory consists of a small-scale surface area of about 40000 square meters, essentially flat, characterized by vegetation types similar to those of the Mediterranean scrub. It is displayed as a grid of 2-meter-long side cells created in the form of a matrix in MATLAB® environment. Different matrices are used to characterize the parameters involved in fire: wind velocity and direction, vegetation density, moisture content, and type (grass and shrubs). All the parameters employed in the CA model are included in Table 3. Random matrices for the vegetation density and typology covering the entire cell grid are generated.

**Table 3. Values for CA model**

The probability values related to density and type of vegetation have been empirically computed by (Alexandridis *et al.* 2008). Other parameters, as constant fire propagation probability $p_0$, and wind coefficient $c_1$ and $c_2$, have been determined by (Alexandridis *et al.* 2008) by performing multiple simulations with non-linear optimization technique

| **Values for the probability $p_{veg}$ and parameter $p_m$** | | |
|---|---|---|
| | *Grass* | *Shrub* |
| $p_{veg}$ | 0.4 | 0.4 |
| $M_d$ | 0.18 | 0.24 |
| **Values for the probability $p_{den}$** | | |
| *Category* | *Density* | $p_{den}$ |
| | Sparse | -0.4 |
| $p_{den}$ | Normal | 0 |
| | Dense | 0.3 |
| **Operational parameters for CA simulations** | | |
| *Parameter* | *Symbol* | *Value* |
| Spread probability under no wind and flat terrain | $p_0$ | 0.6 |
| Wind parameter 1 | $c_1$ | 0.045 |
| Wind parameter 2 | $c_2$ | 0.131 |
| Moisture parameter | $b$ | 0.111 |

To carry out the simulations, a platform managing 120 drones each carrying 20 liters of extinguishing liquid was selected. The number $m_f$ of linear meters of active fire front that can be extinguished by using the platform is computed by (21). Given both this value and the cell size, the number of cells $n_c$ where drones can spread the liquid is calculated in two different ways: if the front develops diagonally, $n_c$ is obtained by dividing $m_f$ with the length of the cell side $l = 2$ multiplied by a factor equal to $\sqrt{2}$, i.e., applying the formula to calculate the diagonal of a square $\sqrt{2} \, l$; otherwise mf is divided only by $l$.

After deciding the position of the platform (on the south side of the domain in the simulation), we faced the flames with a direct attack on both the head and the flank of the fire, as described in [72]. The state of $n_c$ contiguous cells of the fire front closer to the platform is set equal to 4, i.e., in these cells there is a continuous flow of water that extinguishes the fire. With the intervention of drones, the total fire area varies in different ways depending on fire parameters. Specifically, both types of vegetation adopted in the model lead to the outbreak of a fire with low, but rapidly spreading flames. Furthermore, the higher the wind speed, the faster the front spreads, and the more water is needed to extinguish it, all other factors being equal.

Fig. 5 shows the effects of the platform impact on the evolution of a wildland fire. Simulations indicate that although a platform is not able to completely extinguish the fire in these conditions, it is nevertheless effective in containing its advance. The use of two platforms allows the complete extinction.

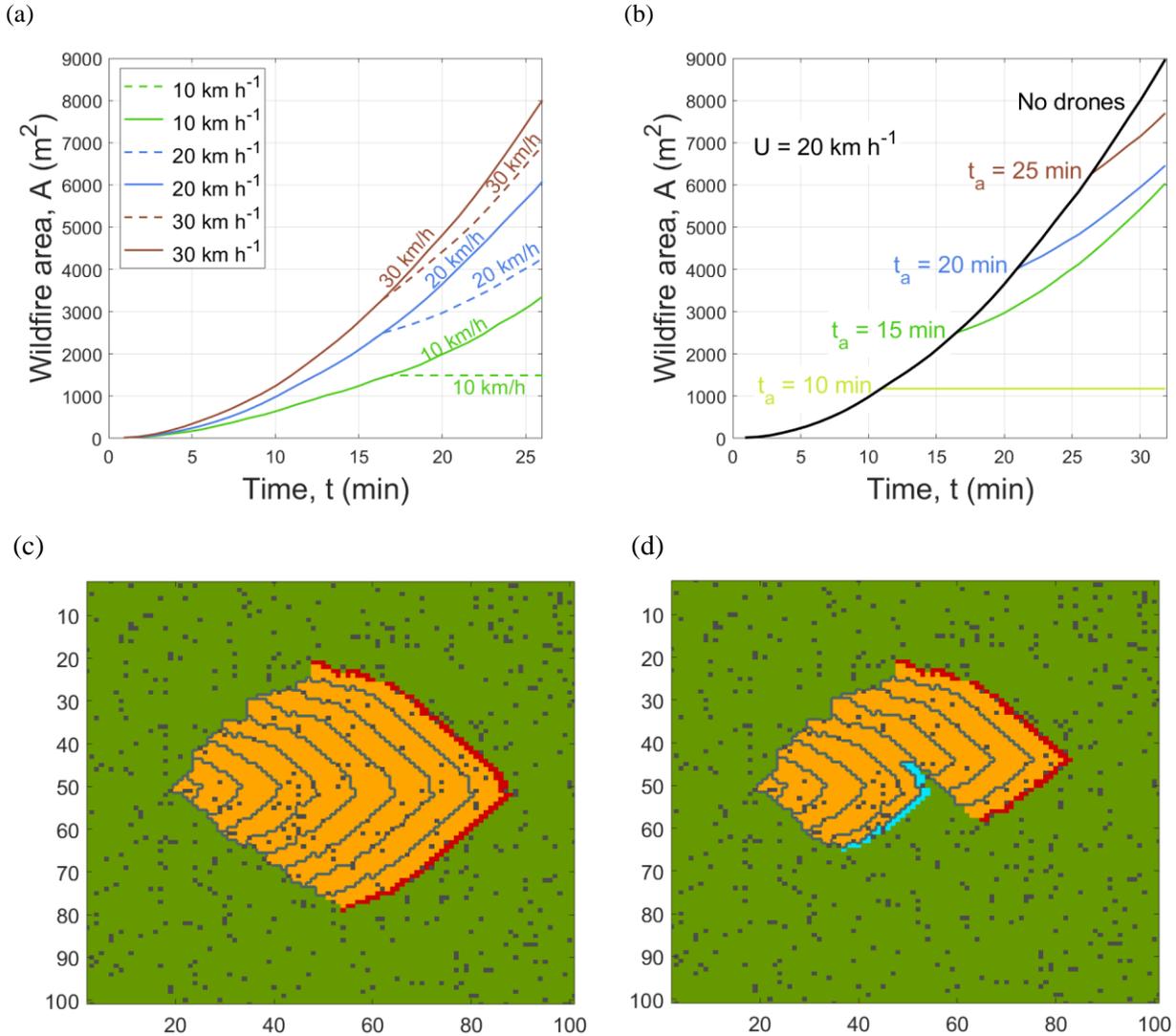

**Fig. 5.** Effects of the proposed firefighting method on the evolution of a wildland fire obtained by CA. (*a*) Variation of the area of a fire in time without any extinguishing intervention (continuous lines) and with the intervention of the firefighting method (dashed line). Drones start to arrive 15 minutes after the fire ignition. The three curves show the evolution of the fire with the same moisture content (18% for grassland, and 24% for shrubland) but with different wind speeds. (*b*) Variation of the area of the fire in time with wind speed fixed at 20 $km\ h^{-1}$, compared with to the spread without any intervention (black line). Drones start the intervention at times $t_a = 10, 15, 20, 25\ min$. (*c, d*) Fire evolution fronts (in grey) without any extinction and with the intervention of drones at $t_a = 15\ min$, respectively. Drones are positioned along the blue line (corresponding to $n_c = 31$ cells). Red cells show the front of the expanding fire.

## Conclusions

We have rigorously estimated the impact of the use of one or more platforms managing a variable number of drones able to spread water or other extinguishing liquid on a wildland fire. On the basis of the critical water flow computed as a function of the main factors involved in the evolution of a fire, we have computed the number of linear meters of active fire front that can be extinguished as these factors vary. We have also tested a fire propagation model to study the evolution of the fire as a result of the containment effect of the platforms. By means of the results of the analyses and graphs carried out in both approaches, the use of a platform for the management of a large number of drones has proven to be a valid method for fighting low-intensity forest fires. As the extinguishing liquid is fractioned into multiple parts, unlike when using aircrafts, future work will investigate a control strategy to decide the part of the fire front where it is preferable to address the action of drones, also using different fire simulation models [73]. Moreover, the system creates the rain effect, i.e., dropping small quantities of firefighting liquid or drizzling it over the fire, instead of spreading it in a concentrated manner. Therefore, it would be interesting to study the rain effect induced by drones in comparison to the impact produced by aircraft carrying the same amount of water. Furthermore, the studies on the mechanics of the platform-drone system will be investigated with regard to technical issues such as the reaction to wind turbulences and high flame temperatures.

**Conflicts of interest**

M.G. is the inventor of the patent "Methods and apparatus for the employment of drones in firefighting activities" [37]. E.A. and P.B. declare they have no conflicts of interest.

**Acknowledgments**

This work received support from Regione Liguria in the context of the European Social Fund 2014-2020 (POR-FSE). We thank archt. Daniele Rossi for figure 1.